\documentclass[conference]{IEEEtran}
\IEEEoverridecommandlockouts
\usepackage{cite}
\usepackage{amsmath,amssymb,amsfonts}
\usepackage{algorithmic}
\usepackage{graphicx}
\usepackage{textcomp}
\usepackage{xcolor}
\usepackage{multirow}
\usepackage{url}
\def\BibTeX{{\rm B\kern-.05em{\sc i\kern-.025em b}\kern-.08em
    T\kern-.1667em\lower.7ex\hbox{E}\kern-.125emX}}
\begin{document}

\title{VisDrone-ReID: A Unified Multi-Category Object Re-Identification Benchmark
}

\author{\IEEEauthorblockN{Asmau Lawal Bilbis, Nathanael L. Baisa, Amudhavel Jayavel}
\IEEEauthorblockA{\textit{School of Computer Science and Informatics} \\
\textit{De montfort University}\\
Leicester, United Kingdom} \\
}

\maketitle

\begin{abstract}
Object re-identification (Re-ID) benchmarks remain predominantly confined to single-category evaluations, limiting the development of unified systems capable of operating across diverse object types. This paper presents VisDrone-ReID, a unified multi-category Re-ID benchmark derived from the VisDrone2019-MOT dataset through a principled conversion pipeline. The pipeline resolves structural incompatibilities between multi-object tracking annotations and Re-ID evaluation requirements, addressing sequence-scoped identity collision, temporal redundancy, and the absence of a principled identity-level train/test split. The resulting benchmark comprises 9,516 identities and 150,016 images across ten object categories — pedestrian, people, bicycle, car, van, truck, tricycle, awning-tricycle, bus, and motor, sourced from 80 unmanned aerial vehicle (UAV) sequences captured across urban environments. A leave-one-out (LOO) evaluation protocol is adopted to provide a comprehensive assessment. Baseline experiments across four representative Re-ID models spanning CNN and transformer architectures establish the first systematic performance reference on this benchmark. Results reveal a consistent Rank-1 to mAP gap across all models, suggesting that VisDrone-ReID presents a genuine retrieval challenge beyond nearest-neighbour matching. Per-category analysis identifies pedestrian as the most challenging category and reveals that vision-language pretraining provides a measurable advantage on semantically distinctive categories. VisDrone-ReID provides a consistent platform for studying multi-category object re-identification under the scale variation, viewpoint challenges, and category diversity characteristic of real-world surveillance scenarios. The source code for dataset generation and evaluation is available at \textcolor{red}{\url{https://github.com/nathanlem1/VisDrone-ReID}}
\end{abstract}


\section{Introduction}
Object Re-identification (Re-ID) is the task of associating a specific object across different times, places, or cameras \cite{BAISA2024104207}. As a foundational 
component of computer vision, Re-ID powers a broad range of applications, including intelligent surveillance, public security, and urban management \cite{ye2021deep}. Real-world surveillance environments are inherently heterogeneous, encompassing a diverse array of entities such as various vehicle types, pedestrians, and cyclists.

While Re-ID models have matured significantly, current benchmarks remain largely siloed into single-category evaluations. Robust, unified multi-category Re-ID systems must demonstrate high performance across multiple object classes simultaneously. Achieving this requires training on diverse, multi-category data. However, dominant benchmarks like Market-1501 \cite{zheng2015scalable} and MSMT17 \cite{wei2018person} focus exclusively on persons, while VeRi-776 \cite{liu2016deep} and VehicleID \cite{guo2016deep} are restricted to vehicles. This category isolation limits the generalisability of models in complex, multi-category scenarios, which is required for different applications such as multi-class multi-object tracking \cite{baisa2025detecttrackuav}.
The VisDrone-MOT (Vision Meets Drones) challenge \cite{zhu2021detection} provides a rich repository of video data captured via Unmanned Aerial Vehicles (UAVs). These sequences present unique challenges, including high traffic density, varied camera motion, and extremely small target resolutions \cite{du2019visdrone}. Furthermore, the significant perspective discrepancy between aerial and ground views has been shown to introduce a substantial domain gap that complicates identity matching \cite{shang2025aerial}. Notably, the dataset features comprehensive multi-category tracking annotations across ten distinct classes (categories), ranging from pedestrians and vehicles to micro-mobility objects. Despite its potential, the VisDrone-MOT structure is designed for temporal continuity and cannot be utilised for Re-ID tasks without a principled conversion pipeline that addresses identity persistence and cross-sequence association.

In this paper, we bridge this gap by converting the VisDrone-MOT dataset into VisDrone-ReID, a unified 10-category benchmark specifically designed for training and evaluating multi-category Re-ID models. The main contributions of this work are as follows:
\begin{itemize}
    \item \textbf{Conversion Pipeline:} We propose a rigorous pipeline 
    and detail the design decisions required to transform raw MOT 
    trajectories into Re-ID identities.
    
    \item \textbf{VisDrone-ReID Benchmark:} We introduce a new 
    large-scale benchmark, providing comprehensive training and testing 
    statistics across ten object categories.
    
    \item \textbf{Evaluation Protocol:} We adopt a leave-one-out (LOO)
    evaluation protocol which provides a comprehensive and fair 
    assessment of retrieval performance.
    
    \item \textbf{Baseline Analysis:} We conduct extensive experiments 
    across four representative Re-ID models 
    \cite{zhou2021learning, luo2019bag, li2023clip, he2021transreid}, 
    establishing a baseline for the community.
\end{itemize}

\section{Related Work}

\subsection{ReID Benchmarks}

The advancement of object re-identification (Re-ID) has historically been driven by the release of specialised benchmarks. These are generally categorised by their target entities and environmental constraints.

\subsubsection{Person Benchmarks} Initial efforts focused on pedestrian retrieval in controlled environments. Market-1501 \cite{zheng2015scalable} introduced a large-scale identity space across six disjoint cameras, defining the standard evaluation protocol. DukeMTMC-reID \cite{ristani2016performance} provided a more challenging environment with eight cameras and significant overlapping views. More recently, MSMT17 \cite{wei2018person} was released to simulate complex surveillance scenarios, featuring 15 cameras and diverse lighting conditions across multiple days. These datasets enabled the development of high-performance architectures, ranging from the widely adopted Bag of Tricks (BoT) \cite{luo2019bag} baseline to modern transformer-based approaches like TransReID \cite{he2021transreid}, which utilizes global context to overcome occlusion. While these datasets are robust, they are restricted to a single object class (pedestrians) and eye-level perspectives.

\subsubsection{Vehicle Benchmarks} To address traffic surveillance, datasets like VeRi-776 \cite{liu2016deep} were developed, offering multi-view annotations for 776 vehicles. VehicleID \cite{guo2016deep} expanded this to over 26,267 vehicles to support large-scale search. Later datasets sought to bridge the gap between different observation platforms; notably, the AGID dataset \cite{shang2025aerial}, which provides a unique setting where vehicle identities are matched across aerial (UAV) and ground-based cameras, highlights the significant domain gap caused by viewpoint variation. Architectural innovations such as OSNet \cite{zhou2019omni}  have shown that omni-scale features are effective for both person and vehicle geometries; however, these models are typically trained on separate single-category datasets due to a lack of unified surveillance data.
\subsubsection{Multi-Category Benchmarks} Multi-category Re-ID is an emerging area that seeks to move beyond single-class silos. A significant recent development is Visual In-Context Prompting (VICP) \cite{huang2025vicp}, which introduces the ShopID10K dataset. ShopID10K consists of 10K instances and 45K images across 34 daily-life categories, collected from online product review platforms, where images reflect natural user-captured variations in lighting, pose, and background. While VICP demonstrates strong zero-shot capabilities, this environment lacks the specific challenges of urban surveillance, such as extreme scale variation, high-altitude viewpoints, and motion blur, that are prevalent in drone-captured footage. Furthermore, while modern architectures like OSNet  \cite{zhou2019omni} and TransReID\cite{he2021transreid} have proven effective across both person and vehicle domains, their evaluation typically relies on separate single-category benchmarks, without a systematic unified protocol for multi-category Re-ID.

\subsection{MOT-to-ReID Synthesis}
Beyond traditionally curated benchmarks, the utilization of Multi-Object Tracking (MOT) trajectories offers an efficient strategy for synthesising Re-ID datasets, since appearance-based re-identification is itself established as a core component within both disjoint and end-to-end MOT pipelines \cite{Baisa2027towards}. This approach leverages the temporal continuity and dense annotations of tracking data to generate robust identity galleries. The VisDrone2019-MOT dataset \cite{zhu2021detection} is particularly suitable for this synthesis due to its high-resolution UAV footage and multi-category annotations.
A primary precedent for this methodology is CityFlow \cite{tang2019cityflow}, a multi-camera vehicle tracking benchmark whose ReID subset demonstrates how tracking trajectories can be repurposed for long-term re-identification. However, this dataset remains limited to the vehicle category and ground-level intersections. Our work applies this synthesis strategy to the drone domain, utilizing the diverse trajectories within VisDrone-MOT to develop a multi-category Re-ID benchmark that includes both pedestrians and various vehicle types from an aerial perspective.
In summary, while Re-ID benchmarks have achieved significant maturity in the person and vehicle domains, they remain largely confined to single-category evaluations and ground-level perspectives. Our work addresses this by transforming the VisDrone tracking trajectories into a dedicated Re-ID benchmark, providing a consistent platform for studying multi-category re-identification in UAV-based surveillance.

\section{VisDrone-ReID Dataset}
The VisDrone2019-MOT benchmark \cite{zhu2021detection}, collected 
by the AISKYEYE team at Tianjin University, China, serves as the source data for this work. The MOT subset provides dense multi-object tracking annotations across ten object categories: pedestrian, people, bicycle, car, van, truck, tricycle, awning-tricycle, bus, and motor, captured from UAVs over urban roads and intersections under varying lighting, weather, and 
altitude conditions.

Despite this richness, the VisDrone-MOT structure is fundamentally incompatible with ReID evaluation in its raw form. Four specific properties prevent direct use. First, object identities are sequence-scoped integers that restart from zero in every sequence, meaning the same integer in two different sequences refers to entirely different physical objects. Second, the dataset contains no quality filter, so many tracklets consist of very few frames, providing insufficient appearance variation for identity learning. Third, uniform frame-by-frame annotation means that naive crop extraction produces highly redundant samples that would artificially inflate evaluation scores. Fourth, there is no principled identity-level train/test split, so without a deliberate conversion strategy, identity leakage between splits cannot be prevented.
These properties collectively motivate the conversion pipeline described below, which addresses each limitation through explicit design decisions to produce a ReID-compatible benchmark.

\subsection{Conversion Pipeline}
\subsubsection{Sequence Assignment}
The 56 annotated training sequences and 7 validation sequences of VisDrone2019-MOT are assigned to the ReID training partition, while 17 sequences from the test-dev subset are held out for the ReID test partition. Sequence-level separation ensures no identity overlap between training and test, since each identity is sequence-local by construction in the source data. The validation sequences are merged into the training partition because their ground-truth annotations are available and suitable for ReID training, maximising the number of labelled identities for model development.

\subsubsection{Global ID Mapping}
VisDrone-MOT assigns object identities as local integers that restart from zero in every sequence, meaning the same integer in two different sequences refers to entirely different physical objects. To resolve this, a deterministic global ID mapping is applied before any filtering. Each unique (\texttt{sequence\_name}, \texttt{local\_track\_id}) pair is assigned a globally unique integer ID via an incremental counter. The first encounter of a pair registers a new global ID; subsequent encounters return the existing mapping. This guarantees that every identity in the dataset refers to exactly one physical object, regardless of which sequence it originated from, producing a collision-free global ID space across all 80 sequences.

\subsubsection{Frame Sampling}
Every 10th frame is sampled from each tracklet. Consecutive frames at full annotation rate exhibit minimal viewpoint or appearance change, producing highly redundant crops that would artificially inflate retrieval scores. Sampling at a fixed interval of 10 frames reduces this redundancy while preserving sufficient appearance diversity across each tracklet for meaningful identity matching.

\subsubsection{Quality Filtering}
Two filtering thresholds are applied. Bounding boxes with an area smaller than 50 pixels are discarded at crop extraction time, as objects at UAV altitude produce crops with insufficient texture detail for appearance-based matching. After all frames are processed, identities with fewer than two images are removed from both partitions, ensuring adequate appearance variation for identity learning and sufficient gallery matches for evaluation. Under the minimum image threshold, 417 training identities and 143 test identities were removed. The smallest retained bounding box area is 77 pixels in the training partition and 60 pixels in the test partition, both above the 50-pixel threshold.

\subsubsection{Category Scope}
All ten VisDrone object categories are retained: pedestrian, people, bicycle, car, van, truck, tricycle, awning-tricycle, bus, and motor. The people and pedestrian categories are maintained as distinct classes following the original VisDrone annotation scheme, in which pedestrian denotes upright walking individuals while people encompass seated, crouching, or densely grouped persons. Merging these categories would require subjective decisions about borderline cases and would reduce the benchmark's fidelity to the source annotations.

\subsubsection{Naming Convention and Directory Structure}
Crops are saved in a flat per-identity directory structure. Each identity is assigned a dedicated folder named by its six-digit global ID. Within each folder, crops are named according to the format 
\texttt{\{id:06d\}\_\{class\}\_\{index:04d\}.jpg}, where 
\texttt{id} is the six-digit global identity ID, \texttt{class} 
is the object category name, and \texttt{index} is the sequential four-digit number. This embeds the category label directly in the filename, making the object category readable without requiring a separate annotation file and simplifying dataset loading for standard ReID frameworks.

\subsection{Dataset Statistics}
Table~\ref{tab:benchmarks} presents an overview of VisDrone-ReID alongside established single-category ReID benchmarks. VisDrone-ReID contains 9,516 identities (IDs) and 150,016 images across ten object categories, derived from 80 sequences of aerial drone footage. To the best of our knowledge, it is the largest multi-category ReID benchmark derived from UAV surveillance footage in terms of the number of identities.

\begin{figure}[htbp]
\centerline{\includegraphics[width=\columnwidth]{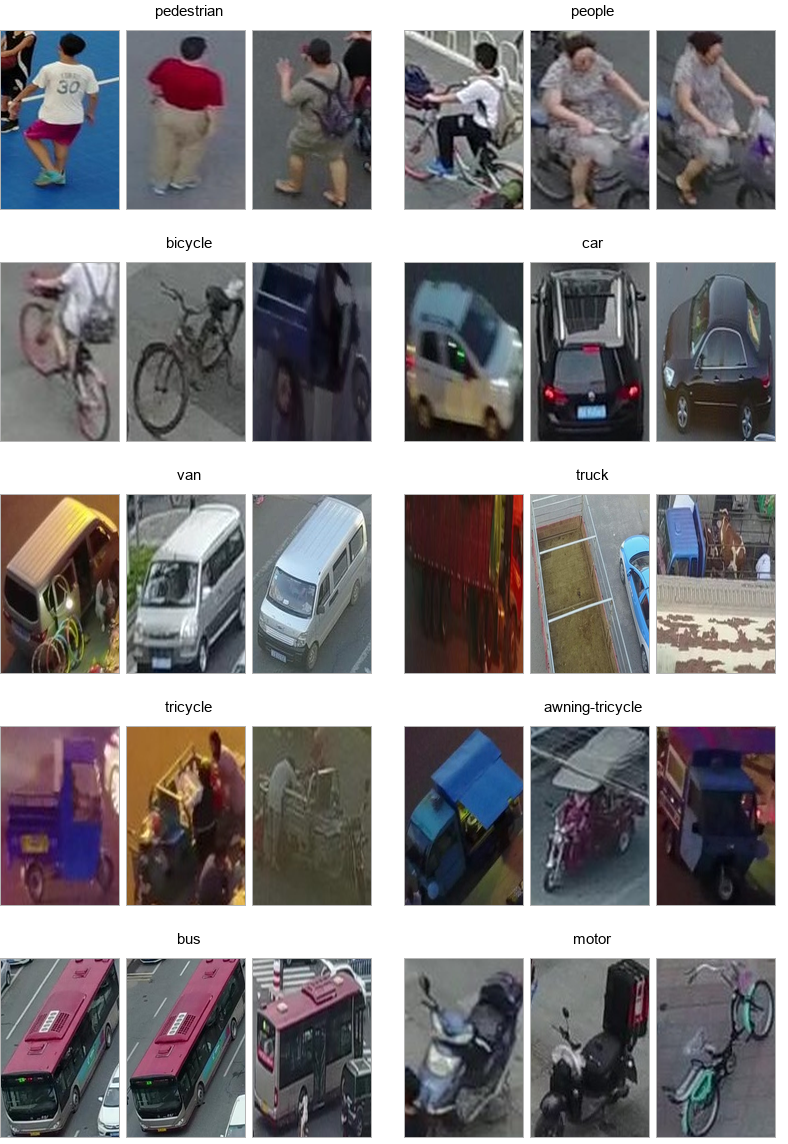}}
\caption{Sample crops from the VisDrone-ReID training partition across all ten object categories. Each row shows three crops of different instances from the same category. All crops are displayed at a uniform size ($180 \times 120$ pixels following $height \times width$ format) for visibility. The figure illustrates the visual diversity within categories, the significant scale variation introduced by UAV altitude, and the challenging appearance conditions, including viewpoint changes, low-light sequences, and partial occlusion that characterise the benchmark.}
\label{fig:crops}
\end{figure}

The training partition comprises 7,486 identities and 119,856 
images drawn from 63 sequences. The test partition comprises 2,030 identities and 30,160 images drawn from 17 sequences. Table~\ref{tab:category_stats} presents the per-category identity 
and image counts across both partitions. The category distribution 
reflects the natural composition of urban aerial surveillance footage, with car and pedestrian accounting for 40.01\% and 24.02\% of training identities, respectively. Less frequent categories, such as bus (0.72\%) and awning-tricycle (1.10\%) are present but under-represented. Fig.~\ref{fig:crops} illustrates representative crops across all ten categories.

Unlike all benchmarks in Table~\ref{tab:benchmarks}, 
VisDrone-ReID does not provide explicit camera IDs, as drone-captured sequences do not correspond to fixed camera positions.

\begin{table}[htbp]
\caption{Comparison of ReID Benchmarks}
\begin{center}
\resizebox{\columnwidth}{!}{%
\begin{tabular}{|l|c|c|c|c|c|c|}
\hline
\textbf{Dataset} & \textbf{IDs} & \textbf{Images} & \textbf{Cats} & \textbf{Cams} & \textbf{View} & \textbf{Source} \\
\hline
Market-1501 \cite{zheng2015scalable}  & 1,501   & 32,668  & 1  & 6   & Ground & Curated \\
MSMT17 \cite{wei2018person}           & 4,101   & 126,441 & 1  & 15  & Ground & Curated \\
VeRi-776 \cite{liu2016deep}           & 776     & 49,357  & 1  & 20  & Ground & Curated \\
CityFlow-ReID \cite{tang2019cityflow} & 666     & 56,277  & 1  & 40  & Ground & MOT-derived \\
\textbf{VisDrone-ReID (ours)}         & \textbf{9,516} & \textbf{150,016} & \textbf{10} & N/A & Aerial & MOT-derived \\
\hline
\end{tabular}}
\label{tab:benchmarks}
\end{center}
\end{table}

\begin{table}[htbp]
\caption{Per-Category Statistics of VisDrone-ReID}
\begin{center}
\resizebox{\columnwidth}{!}{%
\begin{tabular}{|l|c|c|c|c|c|c|}
\hline
\multirow{2}{*}{\textbf{Category}} & 
\multicolumn{3}{c|}{\textbf{Train}} & 
\multicolumn{3}{c|}{\textbf{Test}} \\
\cline{2-7}
& \textbf{IDs} & \textbf{Images} & \textbf{\%IDs} & 
  \textbf{IDs} & \textbf{Images} & \textbf{\%IDs} \\
\hline
Pedestrian      & 1,798 & 26,073 & 24.02\% & 681  & 10,379 & 33.55\% \\
People          & 841   & 10,956 & 11.23\% & 233  & 3,325  & 11.48\% \\
Bicycle         & 315   & 4,538  & 4.21\%  & 23   & 347    & 1.13\%  \\
Car             & 2,995 & 52,952 & 40.01\% & 678  & 9,802  & 33.40\% \\
Van             & 286   & 5,305  & 3.82\%  & 52   & 1,104  & 2.56\%  \\
Truck           & 147   & 3,146  & 1.96\%  & 62   & 1,356  & 3.05\%  \\
Tricycle        & 196   & 3,133  & 2.62\%  & 43   & 669    & 2.12\%  \\
Awning-tricycle & 82    & 1,448  & 1.10\%  & 27   & 429    & 1.33\%  \\
Bus             & 54    & 978    & 0.72\%  & 11   & 152    & 0.54\%  \\
Motor           & 772   & 11,327 & 10.31\% & 220  & 2,597  & 10.84\% \\
\hline
\textbf{Total}  & \textbf{7,486} & \textbf{119,856} & \textbf{100\%} & 
                  \textbf{2,030} & \textbf{30,160}  & \textbf{100\%} \\
\hline
\end{tabular}}
\label{tab:category_stats}
\end{center}
\end{table}

\section{Evaluation Protocol}
\subsection{Leave-One-Out Protocol}
Standard ReID protocols rely on camera IDs to exclude same-camera same-identity pairs from evaluation~\cite{zheng2015scalable}, but VisDrone-ReID provides no camera metadata, as each sequence is filmed by a single moving UAV platform. In this case the most convenient evaluation protocol is leave-one-out (LOO). For each image in the test set, that image serves as the query and all remaining test images constitute the gallery. Correct matches are all gallery images sharing the same identity label as the query. An identity with $N$ images contributes $N$ queries, 
yielding 30,160 queries in total across 2,030 identities (with a total of 30,159 gallery images i.e. $N$-1). This protocol provides a comprehensive assessment as every image participates as a query rather than a single representative 
image per identity. Evaluation reports Rank-1, Rank-5, and Rank-10 accuracy and mean average precision (mAP), following standard ReID evaluation convention~\cite{zheng2015scalable}.

\section{Experiments}
\subsection{Models}
Four representative Re-ID models spanning distinct architectural families are evaluated as baselines. OSNet-IBN \cite{zhou2021learning} is a lightweight CNN with omni-scale feature learning and instance-batch normalisation (IBN) for cross-domain robustness, producing 512-dimensional embeddings. Bag of Tricks (BoT) ResNet50 \cite{luo2019bag} is a standard CNN baseline employing a ResNet50 backbone with batch normalization neck (BNNeck), label smoothing, and combined triplet and center loss, producing 2048-dimensional embeddings. TransReID \cite{he2021transreid} is a pure transformer-based model employing a ViT-B/16 backbone with the Jigsaw Patch Module (JPM) for local feature learning, producing 3840-dimensional pre-BNNeck embeddings. CLIP-ReID \cite{li2023clip} adapts a CLIP \cite{radford2021learning} ViT-B/16 image encoder for ReID through learnable prompt tokens and a dual BNNeck architecture, producing 1280-dimensional pre-BNNeck embeddings.  Table~\ref{tab:models} summarises the architectural properties of all four models.

\begin{table}[htbp]
\caption{Model Architecture Overview}
\begin{center}
\resizebox{\columnwidth}{!}{%
\begin{tabular}{|l|c|c|c|c|}
\hline
\textbf{Model} & \textbf{Backbone} & \textbf{Params} & \textbf{FLOPs} & \textbf{Embed Dim} \\
\hline
OSNet-IBN \cite{zhou2021learning}     & OSNet       & 2.19M  & 0.98G  & 512  \\
BoT ResNet50 \cite{luo2019bag}                    & ResNet50    & 38.84M & 4.08G  & 2048 \\
TransReID \cite{he2021transreid}                  & ViT-B/16+JPM & 86.52M & 10.14G & 3840 \\
CLIP-ReID \cite{li2023clip}                       & ViT-B/16    & 86.14M & 11.35G & 1280 \\
\hline
\end{tabular}}
\label{tab:models}
\end{center}
\end{table}

\subsection{Training Setup}
All models are trained on the VisDrone-ReID training partition of 7,486 identities and 119,856 images with an input resolution of $256 \times 128$ (height $\times$ width). Data augmentation includes random horizontal flipping and random erasing \cite{zhong2020random} for all models. All four models are trained using the default configurations from their respective 
original codebases.

\subsection{Zero-Shot Baseline}
To validate that VisDrone-ReID provides domain-specific supervision beyond what general Re-ID pretraining captures, OSNet-IBN is evaluated on the test partition without any VisDrone training. Table~\ref{tab:zeroshot} presents results using weights of OSNet-IBN pretrained on Market-1501 using the LOO protocol.

\begin{table}[htbp]
\caption{Zero-Shot Baseline --- OSNet-IBN with Market-1501 
Pretrained Weights evaluated on VisDrone-ReID dataset}
\begin{center}
\resizebox{\columnwidth}{!}{%
\begin{tabular}{|l|c|c|c|c|c|}
\hline
\textbf{Weights}  & \textbf{mAP} & 
\textbf{R-1} & \textbf{R-5} & \textbf{R-10} \\
\hline
OSNet trained on Market-1501   & 41.41\% & 75.41\% & 88.51\% & 91.89\% \\
OSNet trained on VisDrone-ReID  & 66.60\% & 90.93\% & 96.79\% & 97.79\% \\
\hline
\end{tabular}}
\label{tab:zeroshot}
\end{center}
\end{table}

Without domain-specific training, OSNet-IBN achieves 75.41\% Rank-1 but only 41.41\% mAP, suggesting that nearest-neighbour retrieval can succeed even when the feature space is poorly calibrated for the target domain.

\subsection{Overall Results}
Table~\ref{tab:overall} presents overall retrieval performance under the LOO protocol across all four models. Results are reported on the test partition of 2,030 identities and 30,160 images, yielding 30,160 queries.

\begin{table}[htbp]
\caption{Overall Results --- LOO Protocol}
\begin{center}
\begin{tabular}{|l|c|c|c|c|}
\hline
\textbf{Model} & \textbf{mAP} & \textbf{R-1} & \textbf{R-5} & \textbf{R-10} \\
\hline
OSNet-IBN    & 66.60\% & 90.93\% & 96.79\% & 97.79\% \\
BoT ResNet50 & 59.28\% & 89.61\% & 95.48\% & 96.61\% \\
TransReID    & 71.30\% & 92.22\% & 97.60\% & 98.42\% \\
CLIP-ReID    & 69.13\% & 91.83\% & 97.52\% & 98.36\% \\
\hline
\end{tabular}
\label{tab:overall}
\end{center}
\end{table}

Under the LOO protocol, all four models achieve Rank-1 accuracy above 89\%, yet mAP figures are substantially lower, ranging from 59.28\% to 71.30\%. This consistent Rank-1 to mAP gap indicates that while nearest-neighbour retrieval is frequently correct, exhaustive retrieval of all relevant gallery images remains challenging across all architectures. This gap motivates the use of mAP as the primary evaluation metric for VisDrone-ReID.
TransReID achieves the highest mAP at 71.30\%, followed by CLIP-ReID at 69.13\%, OSNet-IBN at 66.60\%, and BoT at 59.28\%. The two transformer-based models outperform the two CNN-based models on mAP across this diverse multi-category identity space. The zero-shot baseline of 41.41\% mAP further contextualises these results, with domain-specific training on VisDrone-ReID yielding a 25.19-point mAP gain over Market-1501 pretrained weights for OSNet-IBN, confirming that the benchmark provides supervision beyond what general Re-ID pretraining captures.

\subsection{Per-Category Analysis}
Table~\ref{tab:map} and Table~\ref{tab:rank1} present LOO retrieval performance broken down by object category for all four models. During the per-category evaluation, a common gallery consisting of all test images except a query image is used. 

\begin{table}[htbp]
\caption{Per-Category mAP Results --- LOO Protocol}
\begin{center}
\resizebox{\columnwidth}{!}{%
\begin{tabular}{|l|c|c|c|c|c|}
\hline
\textbf{Category} & \textbf{Queries} & \textbf{OSNet} & \textbf{BoT} & \textbf{TransReid} & \textbf{CLIP-ReID} \\
\hline
awning-tricycle & 429    & 83.47\% & 78.99\% & 87.69\% & 85.91\% \\
bicycle         & 347    & 63.26\% & 56.01\% & 66.49\% & 63.45\% \\
bus             & 152    & 88.68\% & 84.48\% & 89.24\% & 92.07\% \\
car             & 9,802  & 73.79\% & 66.31\% & 78.11\% & 75.75\% \\
motor           & 2,597  & 64.52\% & 56.15\% & 69.39\% & 63.73\% \\
pedestrian      & 10,379 & 58.79\% & 51.24\% & 63.86\% & 63.08\% \\
people          & 3,325  & 63.38\% & 55.74\% & 67.40\% & 65.09\% \\
tricycle        & 669    & 69.32\% & 58.95\% & 76.43\% & 70.41\% \\
truck           & 1,356  & 70.29\% & 66.73\% & 75.65\% & 72.51\% \\
van             & 1,104  & 76.17\% & 71.34\% & 81.22\% & 79.37\% \\
\hline
\textbf{Overall} & \textbf{30,160} & \textbf{66.60\%} & \textbf{59.28\%} & \textbf{71.30\%} & \textbf{69.13\%} \\
\hline
\end{tabular}}
\label{tab:map}
\end{center}
\end{table}

\begin{table}[htbp]
\caption{Per-Category Rank-1 Results --- LOO Protocol}
\begin{center}
\resizebox{\columnwidth}{!}{%
\begin{tabular}{|l|c|c|c|c|c|}
\hline
\textbf{Category} & \textbf{Queries} & \textbf{OSNet} & \textbf{BoT} & \textbf{TransReID} & \textbf{CLIP-ReID} \\
\hline
awning-tricycle & 429    & 98.37\% & 97.44\% & 98.37\% & 98.83\% \\
bicycle         & 347    & 89.34\% & 89.05\% & 90.49\% & 91.07\% \\
bus             & 152    & 98.68\% & 98.68\% & 98.03\% & 96.71\% \\
car             & 9,802  & 95.45\% & 94.15\% & 96.53\% & 96.07\% \\
motor           & 2,597  & 89.45\% & 87.99\% & 90.76\% & 88.83\% \\
pedestrian      & 10,379 & 85.87\% & 84.26\% & 87.58\% & 87.85\% \\
people          & 3,325  & 87.85\% & 86.83\% & 88.93\% & 87.91\% \\
tricycle        & 669    & 96.11\% & 93.87\% & 97.61\% & 96.71\% \\
truck           & 1,356  & 97.35\% & 96.53\% & 97.94\% & 97.57\% \\
van             & 1,104  & 96.47\% & 96.47\% & 97.92\% & 97.37\% \\
\hline
\textbf{Overall} & \textbf{30,160} & \textbf{90.93\%} & \textbf{89.61\%} & \textbf{92.22\%} & \textbf{91.83\%} \\
\hline
\end{tabular}}
\label{tab:rank1}
\end{center}
\end{table}

Four observations are drawn from the per-category results. First, pedestrian is consistently the most challenging category across all models on both Rank-1 and mAP. OSNet-IBN achieves 85.87\% Rank-1 but only 58.79\% mAP on 
pedestrian, while BoT drops to 84.26\% Rank-1 and 51.24\% mAP. Pedestrian accounts for 10,379 queries, the largest alongside car. Small object scale at UAV altitude combined with limited within-sequence appearance variation for walking persons make exhaustive retrieval particularly difficult. Second, bus is the only category where CLIP-ReID outperforms TransReID on mAP, achieving 92.07\% versus TransReID's 89.24\%. 
Bus has visually distinctive and semantically consistent appearance characteristics, and CLIP's language-grounded pretraining may capture these semantic regularities more effectively than purely visual CNN features.
Third, tricycle shows the largest absolute mAP spread across models at 17.48 points, ranging from BoT's 58.95\% to TransReID's 76.43\%. Bicycle also shows high sensitivity at 10.48 points, ranging from BoT's 56.01\% to TransReID's 66.49\%. Both categories have relatively small query counts 
and high visual similarity between instances at aerial viewpoints, suggesting that retrieval performance on rare and visually similar categories is particularly sensitive to architectural choice. Fourth, TransReID achieves the highest mAP in nine out of ten categories, with CLIP-ReID leading only on bus. This suggests that TransReID's overall mAP advantage is not driven by a small number of dominant categories but is consistent across the full category distribution.

\section{Discussion}
The category distribution of VisDrone-ReID is heavily skewed, with car and pedestrian accounting for 40.01\% and 24.02\% of training identities respectively, reflecting the natural composition of urban aerial surveillance footage. As a result, overall accuracy metrics are disproportionately influenced by car and pedestrian performance, and per-category analysis is essential for a complete understanding of model behaviour across the full category distribution.
Several categories have very small test identity counts — bus has 11 test identities and bicycle has 23 — and per-category results for these rare categories should be interpreted with caution. Small sample sizes introduce high variance in retrieval metrics, and a single retrieval failure on a rare category has a disproportionate effect on its mAP relative to a failure on a high-frequency category. Future multi-category ReID benchmarks should prioritise increasing identity counts for under-represented categories to enable more statistically reliable per-category evaluation.
A further structural limitation stems from the MOT source data. Each identity's images originate from a single UAV sequence, so appearance variation is limited to within-sequence viewpoint and scale changes driven by UAV motion rather than genuine cross-viewpoint diversity. As a result, standard cross-viewpoint evaluation protocols cannot be applied, and the leave-one-out evaluation protocol adopted in this work, while appropriate, does not replicate the multi-viewpoint retrieval conditions present in purpose-built traditional single-category ReID benchmarks. More broadly, as VisDrone-ReID is derived entirely from the VisDrone2019-MOT dataset, its scene diversity is bounded by that source's UAV platforms, geographic locations, and imaging conditions; extending the benchmark with additional aerial MOT sources to assess cross-dataset generalisation is a valuable direction for future work. Nevertheless, VisDrone-ReID provides a consistent platform for developing and evaluating unified multi-category ReID models, with direct relevance to multi-object tracking systems that require appearance-based association across diverse object categories and object retrieval systems for traffic monitoring in real-world multi-category scenarios.

\section{Conclusion}
This paper presented VisDrone-ReID, a unified multi-category Re-ID benchmark derived from the VisDrone2019-MOT dataset through a principled conversion pipeline. The pipeline addresses the structural incompatibilities between MOT tracking annotations and ReID evaluation requirements, resolving sequence-scoped identity collisions through global ID mapping, reducing temporal redundancy through frame sampling, and enforcing quality thresholds to produce a dataset of 9,516 identities and 150,016 images across ten object categories. Since each identity's images originate from a single UAV sequence, a leave-one-out evaluation protocol was adopted for comprehensive evaluation in this single-viewpoint setting.

Baseline experiments across four representative Re-ID models — OSNet-IBN, BoT ResNet50, CLIP-ReID, and TransReID — establish the first systematic performance reference on this benchmark. Results reveal a consistent Rank-1 to mAP gap across all models, suggesting that VisDrone-ReID presents a genuine retrieval challenge beyond nearest-neighbour matching. Per-category analysis identifies pedestrian as the most challenging category across all models and reveals that vision-language pretraining in CLIP-ReID provides a measurable advantage on semantically distinctive categories such as bus.
VisDrone-ReID addresses a notable gap in existing ReID benchmarks by providing a unified multi-category surveillance dataset from an aerial perspective, enabling systematic evaluation of Re-ID methods under the object scale variation, viewpoint challenges, and category diversity characteristic of UAV-based deployment scenarios.

\bibliographystyle{IEEEtran}
\bibliography{ref}

\end{document}